\documentclass[10pt,twocolumn,letterpaper]{article}

\usepackage[pagenumbers]{cvpr} 

\usepackage{graphicx}
\usepackage{amsmath}
\usepackage{amssymb}
\usepackage{booktabs}
\usepackage{algorithm, algpseudocode}
\usepackage{subcaption}
\newcommand*\samethanks[1][\value{footnote}]{\footnotemark[#1]}

\algrenewcommand\algorithmicrequire{\textbf{Input:}}
\algrenewcommand\algorithmicensure{\textbf{Output:}}

\usepackage[pagebackref,breaklinks,colorlinks]{hyperref}

\usepackage[capitalize]{cleveref}
\crefname{section}{Sec.}{Secs.}
\Crefname{section}{Section}{Sections}
\Crefname{table}{Table}{Tables}
\crefname{table}{Tab.}{Tabs.}

\def\cvprPaperID{7660} 
\def\confName{CVPR}
\def\confYear{2022}

\begin{document}

\title{Gradient Mask: Lateral Inhibition Mechanism Improves Performance in Artificial Neural Networks}

\author{Lei Jiang \thanks{These authors contributed equally to this work.}\\
Lomonosov Moscow State University \\
{\tt\small lei.jiang@hipasus.com}
\and
Yongqing Liu \samethanks[1]\\
Lomonosov Moscow State University\\
{\tt\small liuyongqing2019@gmail.com}
\and
Shihai Xiao\\
Huawei Technologies Co., Ltd.\\
{\tt\small xiaoshihai@huawei.com}
\and
Yansong Chua \thanks{Corresponding author}\\
China Nanhu Academy of Electronics and Information Technology\\
{\tt\small james4424@gmail.com}
}

\maketitle


\begin{abstract}
Lateral inhibitory connections have been observed in the cortex of the biological brain, and has been extensively studied in terms of its role in cognitive functions. However, in the vanilla version of backpropagation in deep learning, all gradients (which can be understood to comprise of both signal and noise gradients) flow through the network during weight updates. This may lead to overfitting. In this work, inspired by biological lateral inhibition, we propose \textbf{Gradient Mask}, which effectively filters out noise gradients in the process of backpropagation. This allows the learned feature information to be more intensively stored in the network while filtering out noisy or unimportant features. Furthermore, we demonstrate analytically how lateral inhibition in artificial neural networks improves the quality of propagated gradients. A new criterion for gradient quality is proposed which can be used as a measure during training of various convolutional neural networks (CNNs). Finally, we conduct several different experiments to study how \textbf{Gradient Mask} improves the performance of the network both quantitatively and qualitatively. Quantitatively, accuracy in the original CNN architecture, accuracy after pruning, and accuracy after adversarial attacks have shown improvements. Qualitatively, the CNN trained using Gradient Mask has developed saliency maps that focus primarily on the object of interest, which is useful for data augmentation and network interpretability.

\end{abstract}
\section{Introduction}
One may largely divide the discussions on the connection between neuroscience and artificial neural networks (ANN) into two different school of thoughts:
ANN has achieved significant results, many even beyond human performance, with minimal contribution from neuroscience and should proceed as such; another being that, there is a wealth of knowledge in neuroscience that ANN or AI in general can learn from.

On the one hand, one may argue that ANNs were inspired by neuroscience, notable examples could be listed as such: inspired by neuronal working mechanism, the early neural network model was developed \cite{mcculloch1943logical,rosenblatt1958perceptron}; inspired by biological memory storage, Hopfield networks were proposed \cite{hopfield1982neural}; the working mechanism of convolutional neural networks (CNNs) is often considered to be similar to the visual receptive field \cite{lecun1989backpropagation}; based on the firing pattern of biological neurons, spiking neural networks (SNNs) were proposed \cite{cao2015spiking,sengupta2019going,pfeiffer2018deep}; the continuous-attractor neural network (CANN) was inspired by neuronal network dynamics \cite{amari1977dynamics,battaglia1998attractor}; the cortical minicolumn influenced the design of CapsuleNet \cite{hinton2011transforming,sabour2017dynamic}; predictive coding mechanism \cite{rao1999predictive} was applied to ANNs in many practical problems \cite{lotter2016deep,wen2018deep,ye2019anopcn}, and work on contrastive predictive coding \cite{oord2018representation} has led to current trend of unsupervised contrastive learning\cite{henaff2020data,he2020momentum,chen2020simple}.
 

On the other hand, researchers have used neuro-scientific findings to corroborate and explain the effectiveness of ANNs: \cite{pozzi2020attention,lillicrap2020backpropagation} argue for the biological plausibility of back-propagation in the biological brain. In \cite{banino2018vector}, researchers identified a number of neurons sharing grid cell activation characteristics in LSTM networks. OpenAI has also discovered multimodal neurons in ANNs in its recent work \cite{radford2021learning, goh2021multimodal}. 

We have referred to several brain-inspired mechanisms that enhance the performance of ANNs, our work falls in this category.

\begin{figure}[htb]
     \centering
     \begin{subfigure}[]{0.45\textwidth}
         \centering
         \includegraphics[width=\textwidth]{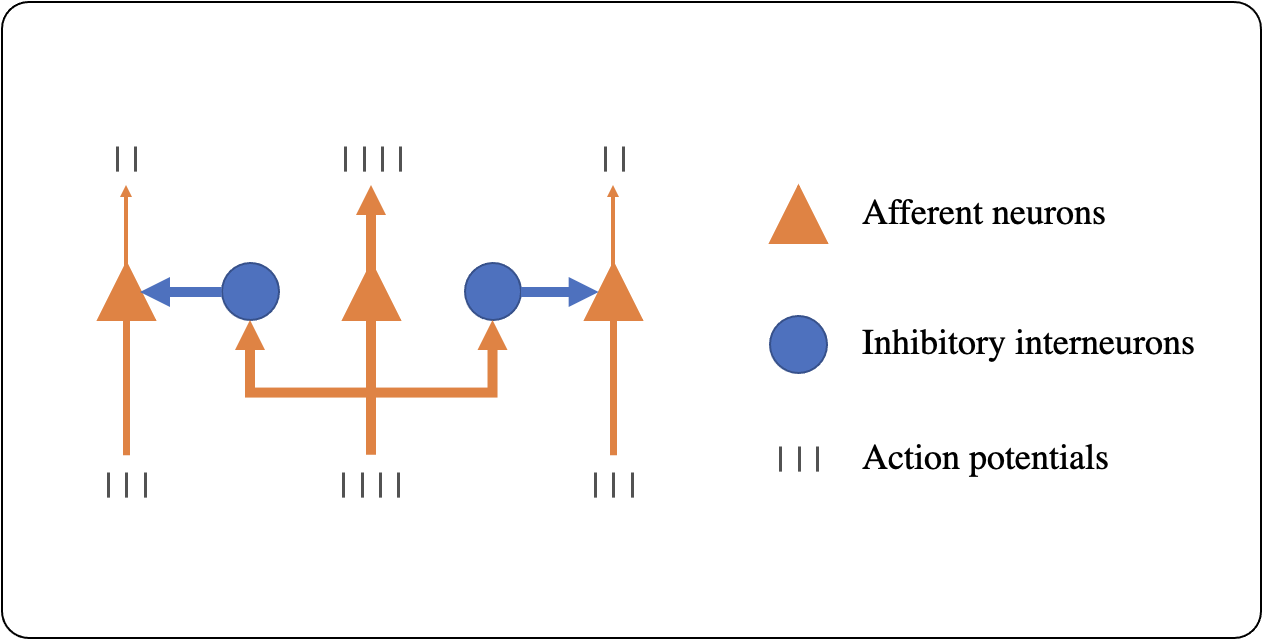}
         \caption{Lateral inhibition process}
         \label{inhibition}
     \end{subfigure}
     \hfill
     \begin{subfigure}[]{0.45\textwidth}
         \centering
         \includegraphics[width=\textwidth]{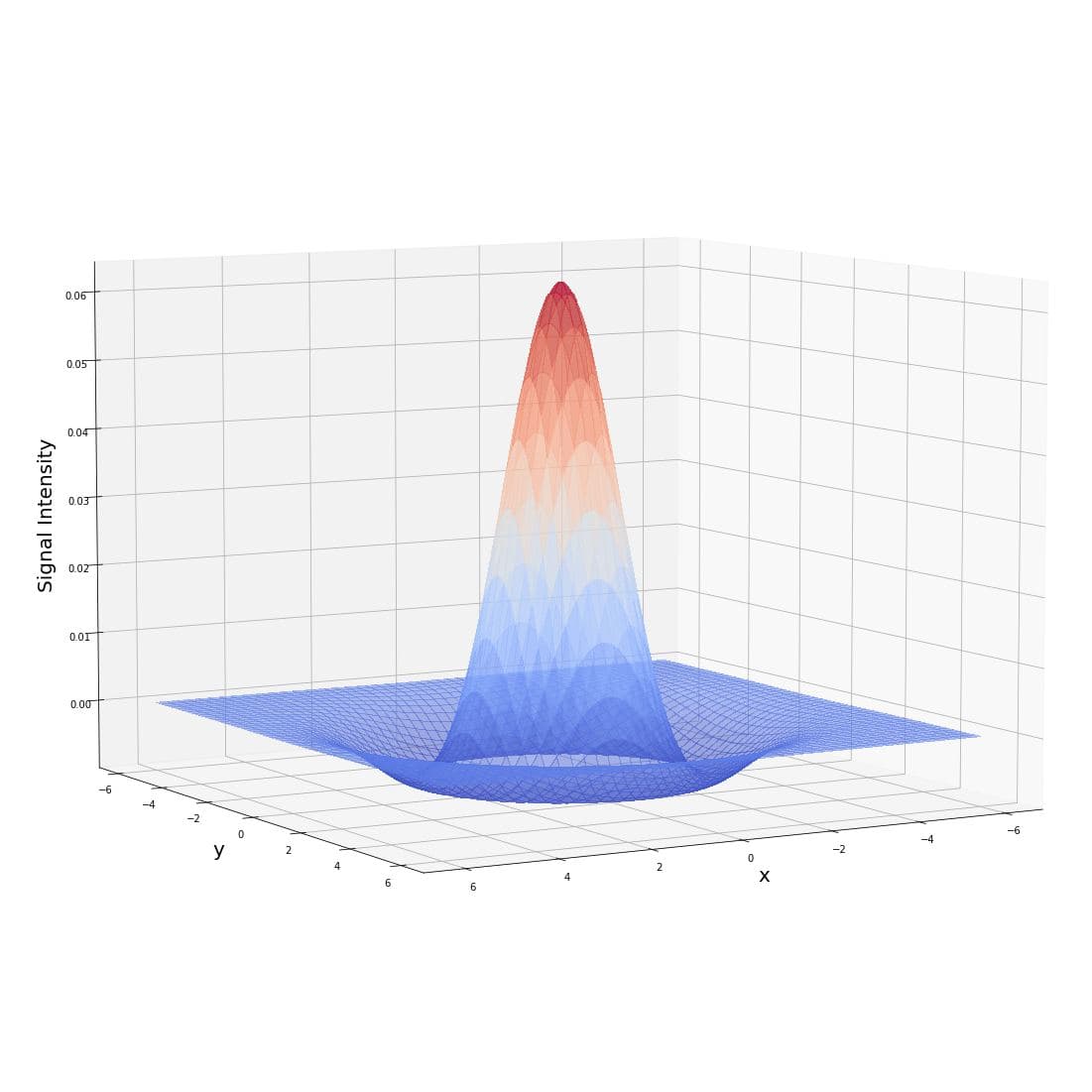}
         \caption{Mexican-hat like distribution}
         \label{mexcian}
     \end{subfigure}
        \caption{In figure \ref{inhibition}, the contrast between strong and weak stimuli is amplified after lateral inhibition; In figure \ref{mexcian}, The Mexican-hat like distribution reflects the effects of LI.}
        \label{fig:two-figures}
\end{figure}

Research in neuroscience has shown that inhibitory circuitry may play an important role in associative LTD (Long-term depression) by ensuring that a postsynaptic cell is inhibited when certain inputs to that cell are active \cite{miller1996synaptic}. And long-lasting changes in the efficacy of synaptic connections between two neurons can involve the making and breaking of synaptic contacts \cite{shoji2007activin}, which is an integral part in the ability of an organism to learn \cite{cooper2005donald}. As a kind of inhibitory mechanism, lateral inhibition (LI) affects the distribution of the attention field by amplifying the contrast between strong and weak stimuli (see Figure \ref{inhibition}). This effect can be described by a Mexican hat function (see Figure \ref{mexcian}). In our model, we use the Laplacian of Gaussian (LoG) operator to approximate this function - highlighting the signal and reducing the noise.

During training of ANN, we hope to inhibit non-critical features to highlight key features. Therefore, we introduce the lateral inhibition (LI) mechanism in ANN training to calculate the importance distribution of the gradients on feature maps, which leads to a new training method Gradient Mask — filtering out the noise gradients, so that only the important gradients could be passed in the backpropagation. The rationality of this method is not only theoretically intuitive (see Section \ref{section gfs}) but also demonstrated in several experiments (see Section \ref{section saliency} and \ref{section gsnr}). 

The effectiveness of the Gradient Mask method is reflected in the following: Firstly, better classification accuracy is obtained by training the network with Gradient Mask (see Section \ref{section training}). Secondly, the training in which the weights are updated by the most relevant gradients leads to a more robust network. On one hand, this method reduces the impact of the unimportant features (as demonstrated by adversarial experiments, see Section \ref{section adversarial}), on the other hand, the signals are packed into a smaller subnet which is sparser but more decisive for learning. (as demonstrated by resistance to pruning, see Section \ref{section pruning}). In addition,  a new data enhancement method by Gradient Mask is proposed, in which we detect the background of the image in real time, and generate new training data by some operations (Gaussian blur, etc.) (see Section \ref{section data-enhancement}).

\section{Methods}
During backpropagation, gradients are generated for all elements the feature map, these then work together to update the corresponding weights. However, previous work \cite{lan2019lca} showed that not every gradient is critical for training, so the importance of the gradient needs to be measured. In order to achieve this goal, we implement the LI mechanism with Laplacian of Gaussian (LoG) operator in the backpropagation process of ANN, because the LI mechanism makes the signal distributed like a Mexican hat, and the LoG operator functions similarly. The operator assigns a value to each feature gradient as a measure of its importance, and through theoretical intuition and experiments we show that this measure of importance is reasonable and effective. During training based on this metric Gradient Mask is generated to filter out unimportant feature gradients, so that back propagation is performed in a subnet.

\subsection{Gradient Mask}
For convolutional layer $l$ with feature maps of width $u$ and height $v$, we divide all the feature maps into $K$ sets evenly. For each set, in order to reduce the computational complexity, the gradients at the same coordinate are composed into a vector called \textit{minicolumn}, which constitutes a fundamental computational unit of the cerebral cortex \cite{szentagothai1975module,swindale1990cerebral,van1993some,mountcastle1997columnar,buxhoeveden2002minicolumn,buxhoeveden2002minicolumn}. We denote the minicolumn on the coordinates $(i, j) $ of the $k$-th set as $M_{ij}^l(k)$. We calculate its $l_2$ norm $||M_{ij}^l(k)||_2$ to represent the magnitude of the gradients in this minicolumn. 

Then for each $k$ we apply the $LoG$ operator to the matrix composed of $||M_{ij}^l(k)||_2$ for all $0\leq i\leq u, 0 \leq j\leq v$. This process is done with $LoG$ convolution kernel,
\begin{align}
    LoG(x,y) 
    &= \frac{\partial^2 G_{\sigma}(x,y)}{\partial x^2}+\frac{\partial^2 G_{\sigma}(x,y)}{\partial y^2}\\
    &= -\frac{1}{\pi \sigma^4}[1-\frac{x^2+y^2}{2\sigma^2}]e^{-\frac{x^2+y^2}{2\sigma^2}}
\end{align}
where $(x,y)$ refers to the coordinates of the $LoG$ convolution kernel and $G_\sigma (x,y)$ is the Gaussian convolution with standard deviation $\sigma$. Let $\delta^l_{ij}(k)$ denote the result of the $LoG$ convolution on the corresponding part. 

By setting the threshold to $\epsilon$, we can define the set of coordinates on which the gradient is not important,
\begin{equation}
    A^l(k) = \{(i,j): |\delta^l_{ij}(k)|<\epsilon \} 
\end{equation}
Therefor we can generate the \textit{Gradient Mask}: $Mask^l(k)=[a_{ij}]_{u\times v}$, where
\begin{align}
    a_{ij}=\left\{
    \begin{array}{rcl}
    0      &     &, {(i,j)\in A^l(k)}\\
    1    &   & ,{(i,j)\not\in A^l(k)}\\
    \end{array} \right.
\end{align}

Since gradient mask $Mask^l(k)$ corresponds to the $k$-th set of the feature maps, the filters corresponding to this set of feature maps share the same mask $Mask^l(k)$. During backpropagation, each gradient passes through the gradient mask, i.e. multiplied by the mask in an element-wise fashion, and then continues to propagate. Let $L$ denote the loss, the gradient of the weight $w_{mn}^l$ on a filter may be written as
\begin{align}
    \frac{\partial L}{\partial w^l_{mn}} 
    &= \frac{\partial L}{\partial a^l_{11}}\frac{\partial a^l_{11}}{\partial w^l_{mn}}+ \dots
    +  \frac{\partial L}{\partial a^l_{NN}}\frac{\partial a^l_{NN}}{\partial w^l_{mn}}\\
    &= \underbrace{\sum_{(i,j)\in A^l} \frac{\partial L}{\partial a^l_{ij}}\frac{\partial a^l_{ij}}{\partial w^l_{mn}} }_{inhibited ~part}+ \underbrace{\sum_{(i',j')\not\in A^l }\frac{\partial L}{\partial a^l_{i'j'}}\frac{\partial a^l_{i'j'}}{\partial w^l_{mn}}}_{important~ part}\\
    &= 0 + \underbrace{\sum_{(i',j')\not\in A^l }\frac{\partial L}{\partial a^l_{i'j'}}\frac{\partial a^l_{i'j'}}{\partial w^l_{mn}}}_{important~ part}
\end{align}

The unimportant feature gradients are set to 0; therefore, the weight only changes in the direction indicated by the important feature gradients, which helps to denoise the feature information retained in the weight \cite{allen2020feature}, improving robustness of the network. This is illustrated in experiments which test the performance of our model against adversarial attacks \ref{section adversarial}. The denoising effect is also illustrated in the GSNR (Gradient Signal to Noise Ratio) (see Section \ref{section gsnr}).

\begin{figure}[htb]
  \centering
  \includegraphics[width=0.45\textwidth]{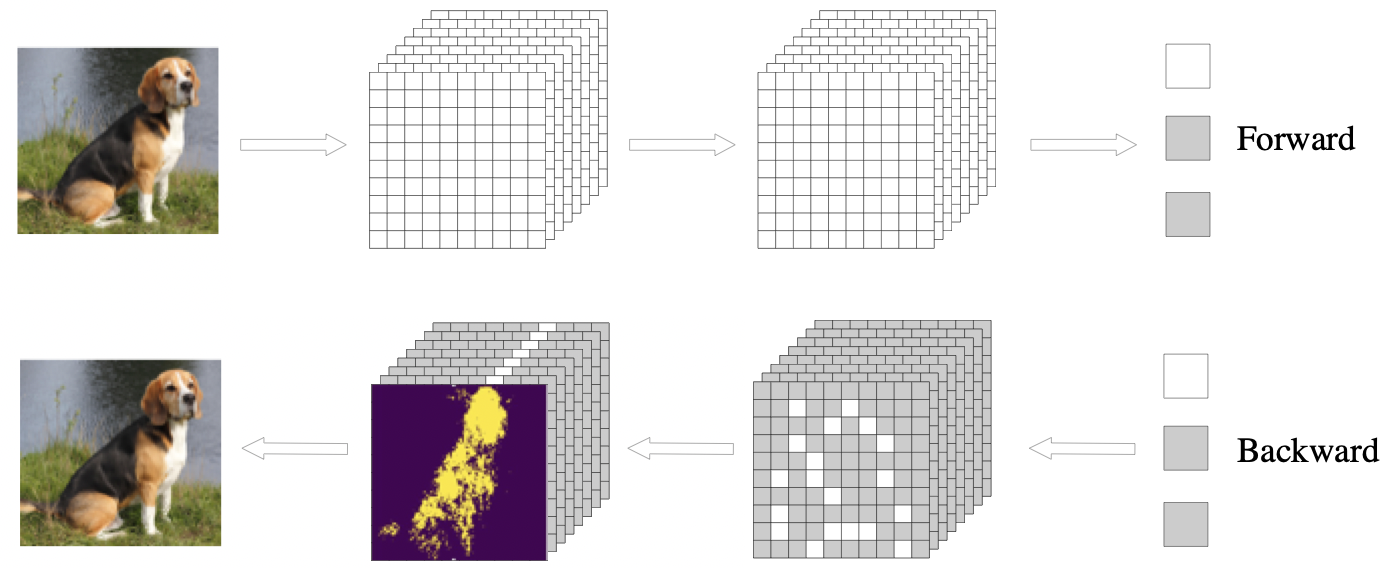}
  \caption{Lateral Inhibition in neural networks. The white square represents the activated neuron while the gray square is inhibited, and the gradient is only flowing through the activated neurons.}
  \label{li-process}
\end{figure}

With gradients masked during back-propagation, given the same accuracy, one can intuitively understand that most features are packed into a smaller subset of useful weights. This is illustrated in the network loss of accuracy during pruning, as discussed in section \ref{section pruning}. Figure \ref{li-process} describes how Gradient Mask is generated and affects the backpropagation. Algorithm \ref{li-method} illustrates the entire process of our method.

\subsection{Gradient Flux Sensitivity} \label{section gfs}
In this section, we provide the theoretical intuition of the Gradient Mask, which is to show the rationality of using $\delta^l_{ij}$ to measure the importance of the feature gradients.

In backpropagation, as shown in equation (5), a weight gradient can be decomposed into multiple terms, and each term corresponds to a coordinate on the feature map. In order to express how significant a point on the feature map is to a weight gradient $g^l_{mn}$, we calculate its laplacian along the coordinates $(i,j)$ on the feature map:
\begin{equation}
    \Delta g_{mn}^l = \frac{\partial^2 g_{mn}^l }{\partial i^2} + \frac{\partial^2 g_{mn}^l }{\partial j^2}
\end{equation}
Physically speaking, it is the divergence of the change of the gradient $g^l_{mn}$ in space, and its absolute value indicates to what extent the point $(i,j)$ is the "source" (positive or negative source) of the gradient $g^l_{mn}$ \cite{katz1979history}. A larger source means that it is more significant and critical for the gradient $g^l_{mn}$. We define it as Gradient Flux Sensitivity $s^l_{mn} = |\Delta g_{mn}^l|$. 

Let $\overline{\frac{\partial L}{\partial a^l_{ij}}}$ be the feature gradient after Gaussian smoothing. So that the gradient flux sensitivity can be rewritten as follows:
\begin{align}
    s_{mn}^l 
    &= \left|\Delta \sum_{i,j} (\overline{\frac{\partial L}{\partial a^l_{ij}}}\frac{\partial a^l_{ij}}{\partial w^l_{mn}})\right| \\
    &= \frac{\partial a^l_{ij}}{\partial w^l_{mn}} \left|\Delta \overline{\frac{\partial L}{\partial a^l_{ij}}}\right|
\end{align}
The coefficient $\frac{\partial a^l_{ij}}{\partial w^l_{mn}}$ is  the  derivative  of  the  activation function  given  the  input  of  the  neuron,  which  can  be  regarded as a constant, and is non-negative since ReLU is the activation function. So the element of the Gradient Mask $\delta^l_{ij} = \left|\Delta \overline{\frac{\partial L}{\partial a^l_{ij}}}\right|$ is positively related to $s_{mn}^l$, therefore $\delta^l_{ij}$ can measure the importance of feature gradients to a weight gradient.

%

\begin{algorithm}{}
  \caption{Generation of Gradient Masks for one sample in layer $l$}
  \label{li-method}
  \begin{algorithmic}[1]
    \Require
      Tensor of gradients in layer $l$, Quantile, Sigma, Kernel size, Number of channels, Number of sets.

    \State Initialize $G^l$ = Tensor of gradients in layer $l$, $q$ = Quantile, $\sigma$ = Sigma, $s$ = Kernel size, $C$ = Number of channels , $K$ = Number of sets. Let $l_c$ be the $c$-th feature map of layer $l$,
    \begin{align}G^l = [\frac{\partial{L}}{\partial{a_{ij}^{l_c}}}]_{C\times u\times v}
    \end{align}
    \State Divide gradients into minicolumns:
    \begin{align}
        M^l_{ij}(k) = [\frac{\partial{L}}{\partial{a_{ij}^{l_{c'}}}}]_{\frac{C}{K}\times 1}
    \end{align}
    and compute $l_2$ norm of each minicolumn: $||M^l_{i,j}(k)||_2$ to get $K$ matrices:
    \begin{align}
        D^l(k)= [||M^l_{ij}(k)||_2]_{u\times v}, 
    \end{align}
    \State Perform $LoG$ convolution on each matrix $D^l(k)$ with given parameters $\sigma$ and $s$: 
    \begin{align}
        \delta_{ij}^l(k) = LoG_{\sigma,s}(D^l(k))
    \end{align}
    \State Calculate threshold value $Q_k^l$ with given parameter quantile $q$: 
    \begin{align}
        Q^l_k=Threshold({[\delta_{ij}^l(k)]_{u\times v},q)}
    \end{align}
    \State Generate Gradient Masks:
    \begin{align}
    	Mask^l(k)=[\mathbb{I}_{|\delta^l_{ij}(k)|>Q_k}]_{u\times v}
    \end{align}
    \State Update the gradients on each feature map $c$ with the Gradient Mask:
    \begin{align}
       \overline{G_c^l} = [\frac{\partial{L}}{\partial{a_{ij}^{l_c}}}]_{u\times v}\circ Mask^l(k)|_{k\in\{k|l_c\in set (k)\}}
    \end{align}
    \State Continue backpropagation with $\overline{G^l}= [\overline{G^l_c}]_{C\times 1}$.
    \end{algorithmic} 
\end{algorithm}


\section{Experiments}
Training with Gradient Mask, our model achieved higher accuracy on ImageNet \cite{deng2009imagenet} and CIFAR-100 \cite{krizhevsky2009learning} datasets. And In order to visualize the effect of the Gradient Mask method, we use the LoG operator to generate saliency maps, and compare them with the saliency maps generated by other methods like Grad-CAM\cite{selvaraju2017grad}, verifying that this method is able to more accurately capture key features. 

The network trained with Gradient Mask has also improved in robustness and generalization ability, which comes from the accumulation of signals and the inhibition of noise:
\begin{itemize}
\item In terms of signal, the pruning experiment shows that the model can be pruned more while maintaining accuracy, indicating that the key gradients are indeed packed into a sparser sub-network;
\item In terms of noise, adversarial experiments show that the noise gradients are less learned by the model.
\item Furthermore, we quantitatively demonstrate the relative enhancement of the signal and the reduction of the noise by calculating the Gradient Signal to Noise Ratio (GSNR) \cite{liu2020understanding}.

\end{itemize}

Finally, applying Lateral Inhibition, we propose a new data enhancement scheme that can detect and blur unimportant parts of the images in real time.

\subsection{Training with Gradient Mask} \label{section training}
To fairly compare the performance of the original network and the network trained with Gradient Mask, we use ResNet\cite{he2016deep} as the experimental framework and apply the same hyperparameters and data enhancement strategy. In order to generate the Gradient Mask, we set additional hyperparameters $\sigma=11$, $K=16$, and the quantile(inhibition rate) at $50\%$. We apply Gradient Mask on the output layer of every bottleneck except the last two bottlenecks, which contain rich semantic information, and too much inhibition would lead to degradation of network performance. 
Although using the same learning rate as the original network is not the optimal choice for the Gradient Masked network, as the latter only updates the weights in its sub-network, but in order to make a fair comparison, we use the same hyperparameters as the original network.
We conducted classification experiments on ImageNet (with 8 Tesla V100 GPUs) and CIFAR-100 (with 1 Tesla V100 GPU) respectively, and the experiments showe that using Gradient Mask improves classification accuracy. Table \ref{resnet} shows the result of ResNet with/without gradient masks in the two datasets.

\begin{table}[htb]
  \centering
  \resizebox{0.45\textwidth}{!}{
  \begin{tabular}{lll}
    \toprule
    Model         & CIFAR-100    &  ImageNet     \\ 
    \midrule
	Normal ResNet-18/50       &  78.2      &     75.5         \\
	Masked ResNet-18/50     &  \textbf{80.26} & \textbf{76.01} \\
    \bottomrule
  \end{tabular}}
  \caption{The top-1 accuracy (\%) of ResNet-18 with/without masks on CIFAR-100 and ResNet-50 with/without mask on ImageNet}
  \label{resnet}
\end{table}

In order to investigate whether LI and minicolumns are crucial for Gradient Mask to work, two experiments are conducted: in the first experiment, after slicing the pixel channels into minicolumns, we perform an $l_2$ norm operation on each minicolumn, and we took the quantile directly from the norm results to decide which minicolumns should be inhibited/activated (without LI); in the second experiment, we do not aggregate neurons as minicolumn, we take the absolute value of each pixel on each feature map, and then perform $LoG$ on the feature map (without minicolumn).

\begin{table}[htb]
  \centering
  \resizebox{0.45\textwidth}{!}{
  \begin{tabular}{llll}
    \toprule
    Model         &  Masked ResNet-18   &  Without LI & Without Minicolumn   \\ 
    \midrule
	Accuracy     &  \textbf{80.26}      &    75.86    &78.97 \\
    \bottomrule
  \end{tabular}}
  
  \caption{The top-1 accuracy (\%) of ResNet-18 with and without LI and Minicolumn on CIFAR-100}
  \label{tab:li-mini}
\end{table}

\begin{figure}[htb]
  \centering
  \includegraphics[width=0.45\textwidth]{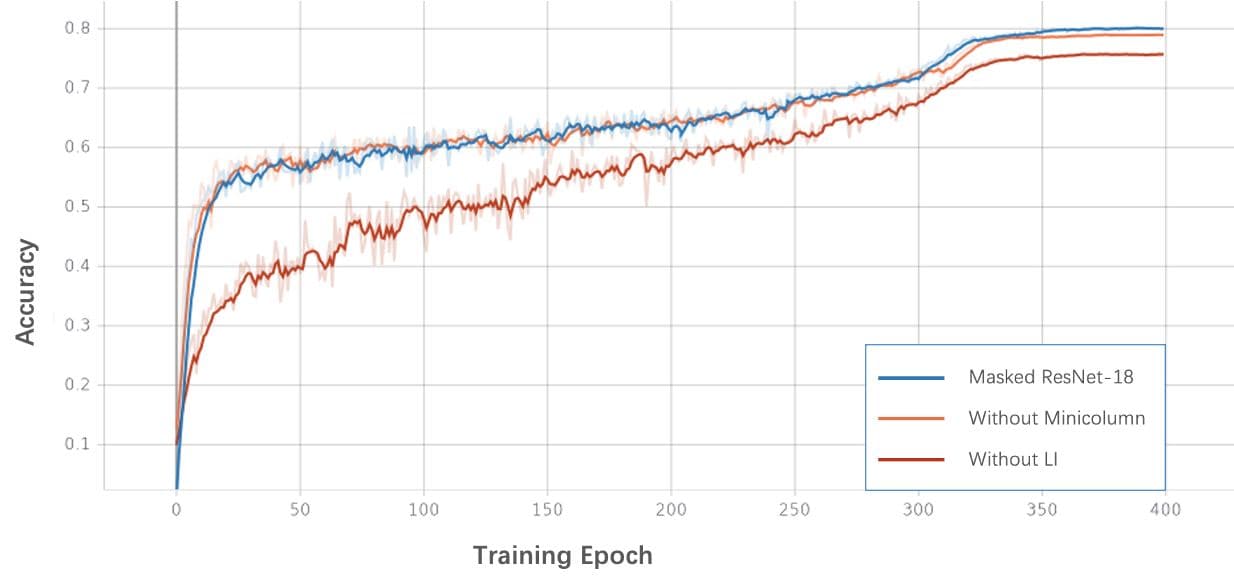}
  \caption{Training with and without LoG and Minicolumn in CIFAR-100 dataset.}
  \label{log_mini} 
\end{figure}

We use ResNet-18 and the same hyperparameters to conduct the experiment in CIFAR-100. The results are shown in Table \ref{tab:li-mini} and Figure \ref{log_mini}, we show that $LoG$ is crucial for the generation of Gradient Mask - \textbf{Without LI} reduces the accuracy significantly. The accuracy of \textbf{Without Minicolumn} is lower than \textbf{Masked ResNet-18}, and its training speed is 3 times slower than \textbf{Masked ResNet-18} (more masks need to be generated).

\subsection{Saliency Map} \label{section saliency}
Saliency map highlights portion of the image that contributes to a classification decision, hence it is an effective tool for the interpretation of convolutional neural networks (CNN). In recent years, various saliency detection methods have been proposed: Guided Backpropagation(GBP) \cite{springenberg2014striving}, Class Activation Mapping(CAM)\cite{zhou2016learning}, Grad-CAM\cite{selvaraju2017grad}, Grad-CAM++\cite{chattopadhay2018grad} and so on. Although some saliency detection methods \cite{springenberg2014striving} are able to generate the saliency map, they are independent of the model and data generation process, and may not best explain the relationship between the inputs and outputs of the model during learning or to debug the model \cite{adebayo2018sanity}.

We used the LI method on a CNN to generate better saliency maps, which we named as \textbf{LI-Map}. Additional experiments were conducted to show that the saliency map generated with LI pass the Cascading Randomization test \cite{adebayo2018sanity} (See \textbf{Appendix A}), confirming that the generated saliency map is helpful for interpretability of the network.

\begin{figure}[htb]
  \centering
  \includegraphics[width=0.45\textwidth]{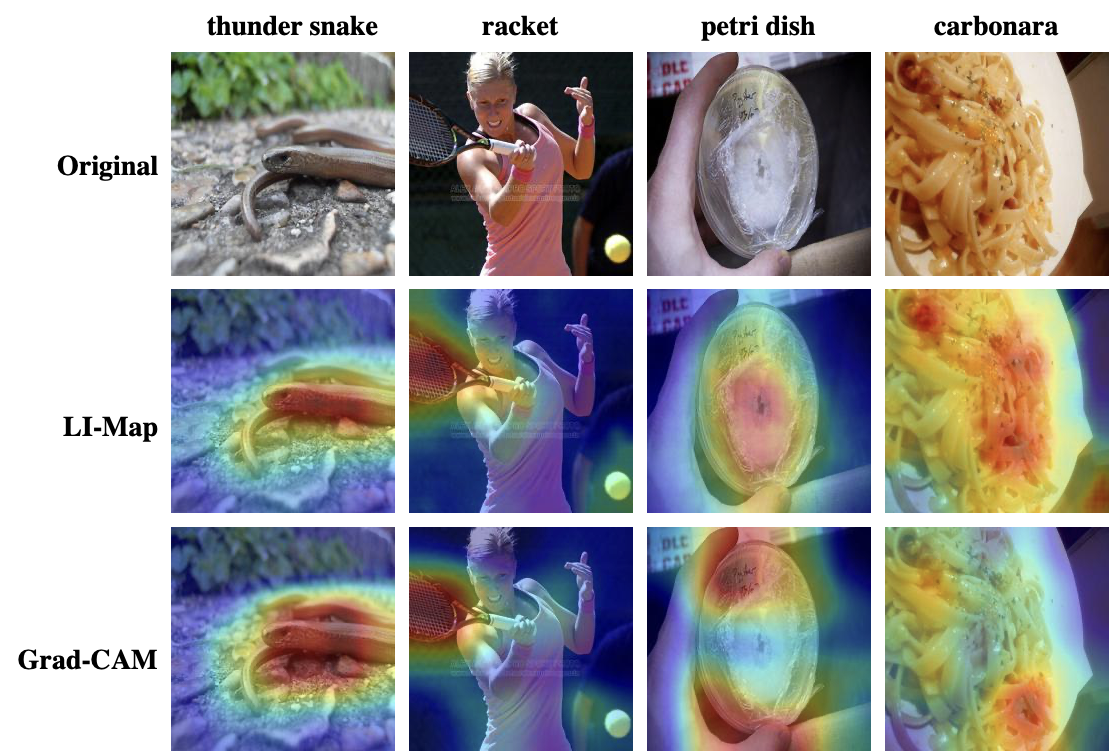}
  \caption{Saliency maps generated by LI-Map and Grad-CAM. Note that the saliency maps generated by LI-Map are more accurately focused on the target objects.  }
  \label{li-map}
\end{figure}

For every activation layer $A^l$, we apply step 1-3 from Algorithm \ref{li-method} to get $\delta^l$. Notice that: 1) Replace loss $L$ with $P$, where $P$ is the prediction of interest; 2) Set $K=1$, i.e. each pixel channel can be seen as one minicolumn. Then resize $\delta^l$ from $u\times v$ to $H \times W$, where $H$ and $W$ are the height and width of the input image:
\begin{align}
    &\delta^l = LoG_{\sigma,s}(D^l) \\
    &\delta^l \in \mathbb{R}^{u \times v}\rightarrow \delta^l \in \mathbb{R}^{H \times W}
\end{align}
Finally, in order to combine information from each layer, we sum the $\delta^l$ of all activation layers to get the saliency map $F$:
\begin{align}
\label{eqn:saliency-f}
    F = \sum_{l=1}^t \delta^l, \quad l \in [1,t]
\end{align}

where $t$ is the number of activation layers. Figure \ref{li-map} shows our results compared with Grad-CAM. We can see that the saliency maps generated by LI-Map are more accurately focused on the target objects.

With the aim of quantitatively comparing the object detection ability of various saliency map methods, we propose a measure based on Intersection over Union (IoU) scores. We select 15\% of pixels with the largest F values (F is calculated by Equation \ref{eqn:saliency-f}) in the saliency map generated by the different methods as the region of interest. The smallest and largest coordinate points of these pixels are used to form a rectangular bounding box and the IoU score is calculated with the real bounding box where the target is located. As shown in Table \ref{iou}, we verify the superiority of our method on the PASCAL VOC 2007 \cite{pascal-voc-2007} and ImageNet datasets.

\begin{table}[htp]
  \centering
  \resizebox{0.45\textwidth}{!}{
  \begin{tabular}{llll}
    \toprule
    Dataset         & Grad-CAM   &   Grad-CAM++ &  LI-Map \\ 
    \midrule
	PASCAL VOC 2007 &  0.44      &      0.45    &  \textbf{0.5} \\
	ImageNet        &  0.46      &      0.46    & \textbf{0.49} \\
    \bottomrule
  \end{tabular}}
  \caption{The IoU results on PASCAL VOC 2007 and ImageNet}
  \label{iou}
\end{table}

In order to further understand the impact of using Gradient Mask, we visualize different saliency maps trained with no mask, up to one mask per image, see \textbf{Appendix B}.

\subsection{Robustness and Generalization performance}


\subsubsection{Network Pruning} \label{section pruning}
\begin{figure}[htb]
\centering
\begin{minipage}[]{0.45\textwidth}
\centering
  \includegraphics[width=\textwidth]{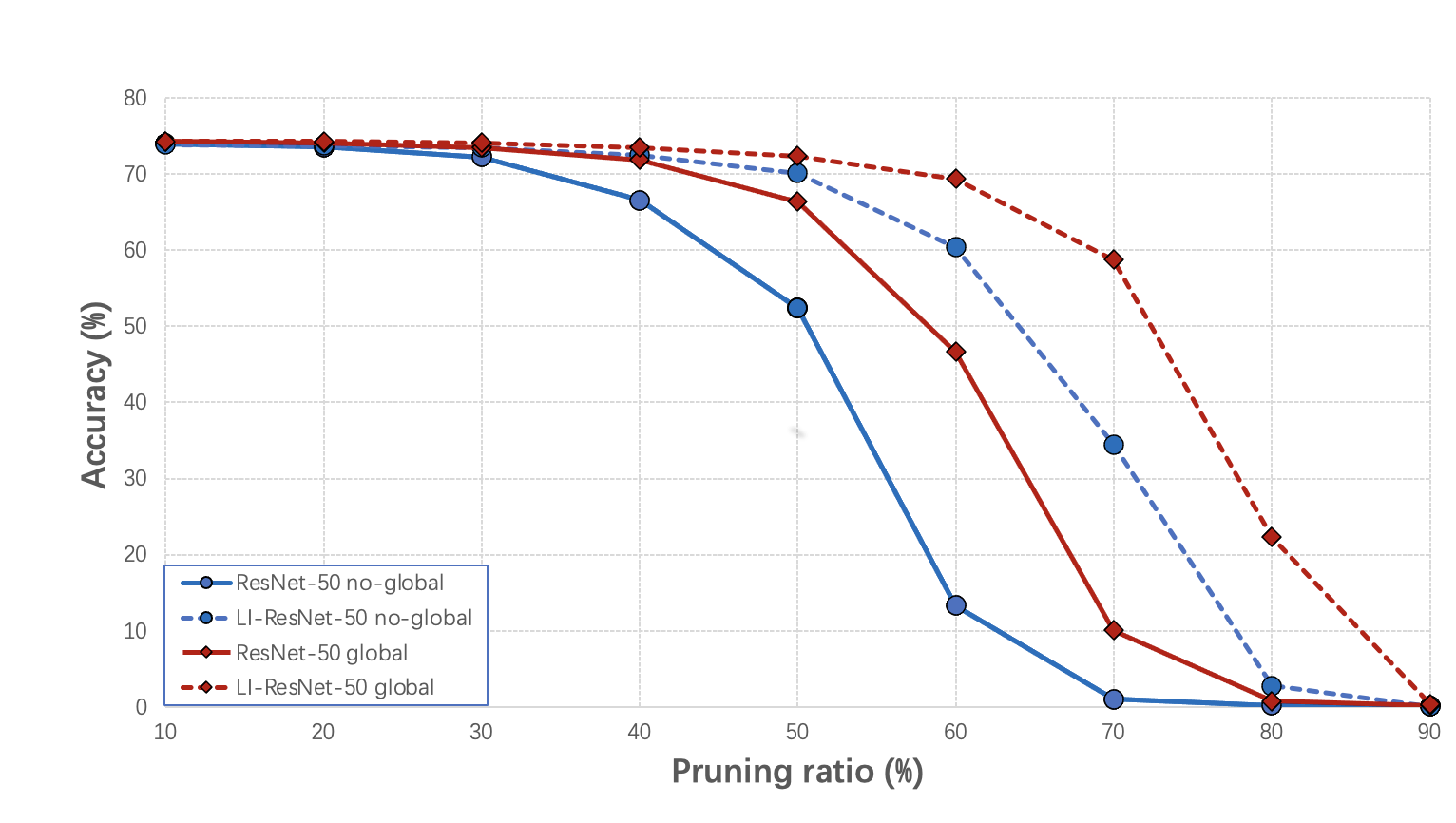}
  \caption{The LI-based network has better performance after pruning}
  \label{pruning}
\end{minipage}
\end{figure}

To confirm that the network trained with Gradient Mask has the ability to intensively learn features into a sparser subnet, we trained a GM-ResNet-50 using Gradient Mask, pruned it with L1 (we pruned a certain percentage of the weights with the smallest absolute value between layers or globally to observe the performance of the network) and compared it with a normally trained ResNet-50.
As shown in Figure \ref{pruning}, the network with Gradient Mask has better performance after pruning.

\subsubsection{Adversarial Attack} \label{section adversarial}
\begin{figure}[htb]
\begin{minipage}[]{0.45\textwidth}
\centering
  \includegraphics[width=\textwidth]{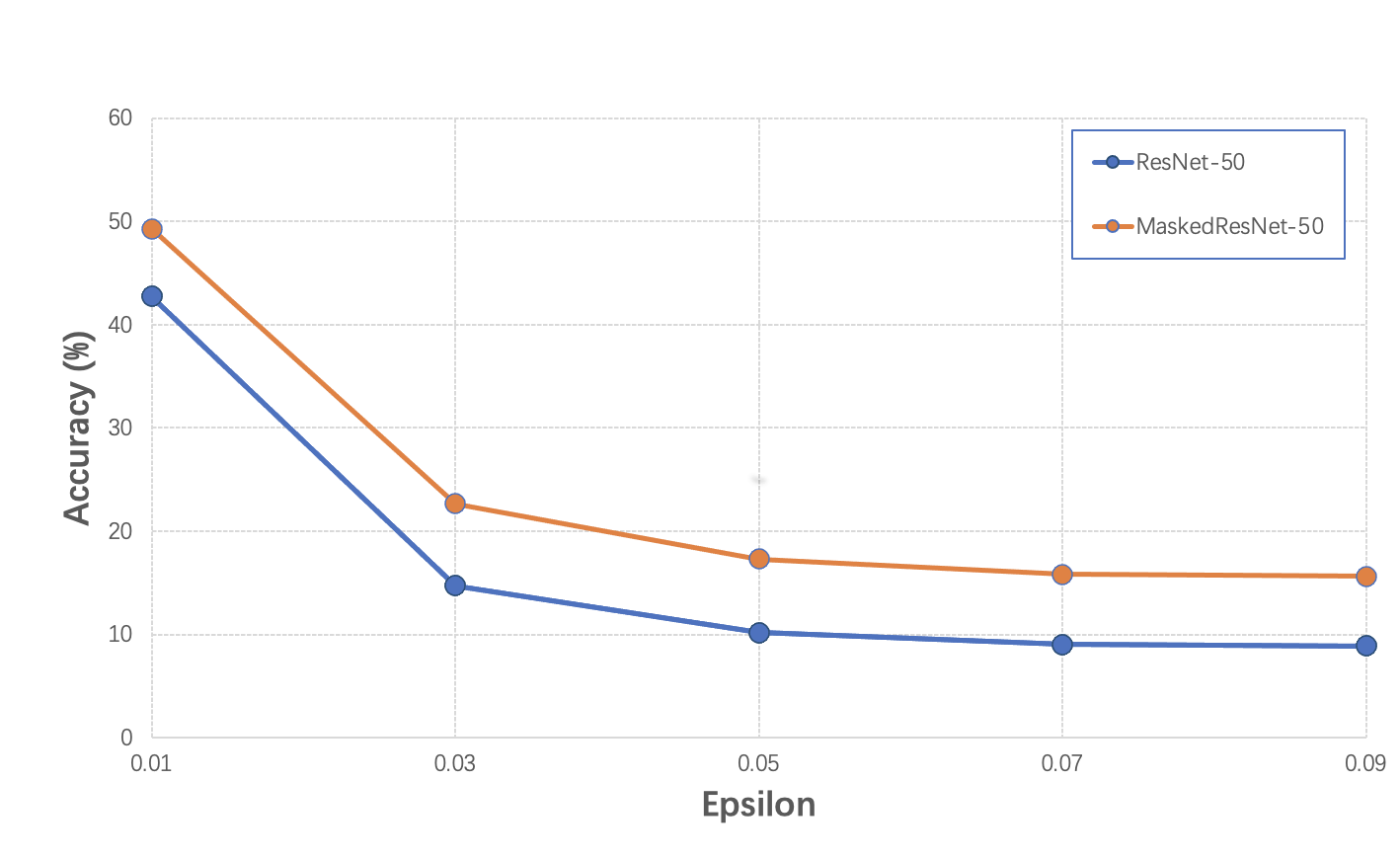}
  \caption{The LI-based network has better capabilities against adversarial attacks compared with ordinary networks.}
  \label{adversarial}
\end{minipage}
\end{figure}

In order to test the robustness of the network trained with Gradient Mask, we conducted adversarial attacks on it. We apply Fast Gradient Sign Attack (FGSM) \cite{goodfellow2014explaining} on both models (GM-ResNet-50 and normal ResNet-50). The dataset comprises of images from ImageNet validation set that can be correctly predicted by both models. As shown in Figure \ref{adversarial}, model trained with Gradient Mask is more robust against adversarial attacks.

\subsubsection{Gradient Mask leads to better GSNR} \label{section gsnr}

\begin{figure}[htb]
\begin{minipage}[]{0.45\textwidth}
\flushright
  \includegraphics[width=\textwidth]{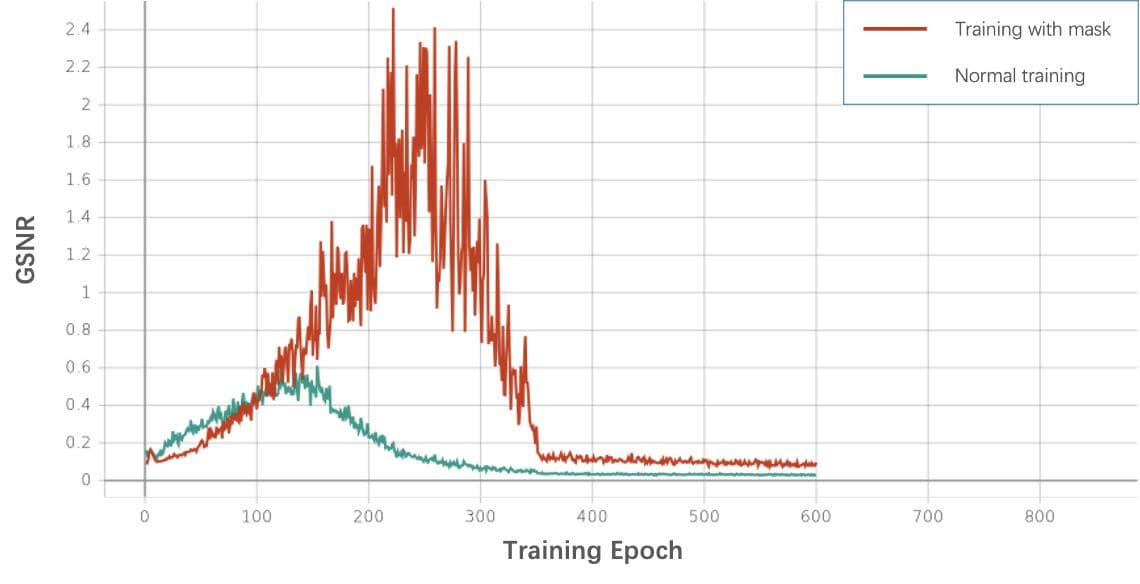}
  \caption{GSNR comparison between normal training and training with Gradient Mask in CIFAR-10}
  \label{gsnr}
\end{minipage}
\end{figure}

In order to test the performance of Gradient Mask in filtering noise gradient, we use the Gradient Signal to Noise Ratio (GSNR) \cite{liu2020understanding} that has been shown to quantify network generalizability. The GSNR of a model parameter $\theta_j$ is defined as:
$$r\left(\theta_{j}\right):=\frac{\tilde{\mathbf{g}}^{2}\left(\theta_{j}\right)}{\rho^{2}\left(\theta_{j}\right)}$$
where $\tilde{\mathbf{g}}\left(\theta_{j}\right)$, $\rho \left(\theta_{j}\right)$ are mean and variance of $\theta_j$ in the iteration, respectively. In our experiment, we compare the GSNR of convolutional layers between normal training and training with Gradient Mask in CIFAR-10 \cite{krizhevsky2009learning}. Figure \ref{gsnr} shows that Gradient Mask can improve GSNR of the model during training, indicating training with Gradient Mask can improve the generalization of the network.

\subsection{Data Enhancement} \label{section data-enhancement}
Data enhancement is an effective method to improve the generalization and robustness of a neural network. Previous works \cite{zhang2020putting, xiao2020noise} demonstrated that context and background will greatly affect the performance of the network and the network can easily overfit to the background during training. As Section \ref{section saliency} shown, LI can help to locate the target of interest in real time, so that we can insert noise in the non-target area (background or other objects) for data enhancement.

\begin{figure}[htbp]
  \centering
  \includegraphics[width=0.45\textwidth]{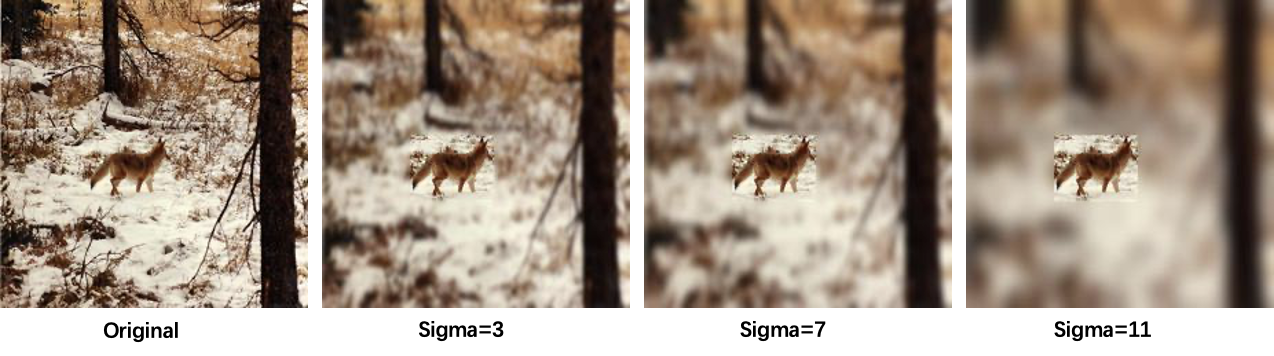}
  \caption{Examples from validation dataset}
  \label{blur_examples}
\end{figure}

\begin{figure}[htbp]
  \centering
  \includegraphics[width=0.45\textwidth]{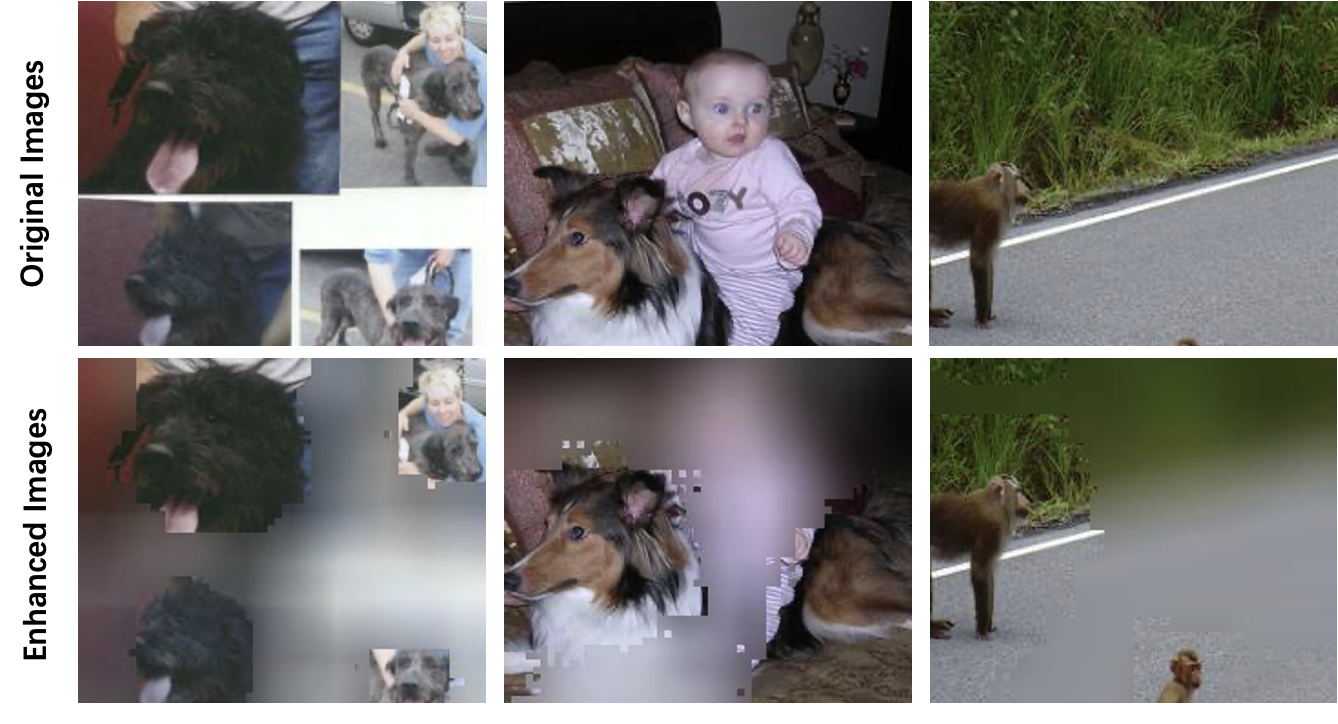}
  \caption{Generation of enhanced images with lateral inhibition.}
  \label{enhanced_images}
\end{figure}

Since no annotation was found for the ImageNet validation set, we selected 100,000 images with bounding box from 1.28 million images as the validation set, and retrained the ResNet-50 on the rest 1.18 million images. Figure \ref{blur_examples} shows the examples from the validation dataset, in which sigma is the blur ratio. Based on the retrained ResNet-50, we continued to train the model using data augmentation for 30 epochs, tested it on 100,000 images with bounding box and Gaussian blur, and compared it with the original pre-trained model. The algorithm is described in Algorithm \ref{enhancing}. Examples of enhanced images are shown in Figure \ref{enhanced_images}, we can observe that the irrelevant information of the target object in the image is well covered. 

\begin{algorithm}{}
  \caption{Data Enhancement with Gradient Mask}
  \label{enhancing}
  \begin{algorithmic}[1]
    \Require
      Input tensor with shape $N \times 3 \times H \times W$, Quantile, Sigma, Kernel size, Set of activation layers
      
    \Ensure  
      Output tensor with shape $N \times 3 \times H \times W$.

    \State Initialize $I$ = Input tensor, $q$ = Quantile, $\sigma$ = Sigma, $s$ = Kernel size, $K$ = Number of Sets = 1, $A=\{A^1,A^2,\ldots,A^t\}$ = Set of activation layers.
    
    \State For every activation layer  $A^l$, $1 \leq l \leq t$ in $A$, apply steps 1-3 in algorithm 1. Then Use linear interpolation to scale $\delta^l$ to input size $(H,W)$.
    \begin{align}
        \delta^l &= LoG_{\sigma,s}(D^l) \\
        \delta^l \in \mathbb{R}^{N \times 1 \times u \times v} &\rightarrow \delta^l \in \mathbb{R}^{N \times 1 \times H \times W}
    \end{align}
    
    \State Summary all $\delta^l$ on second dimession:
    \begin{align}
        F = \sum_{l=1}^t \delta^l, \quad F \in \mathbb{R}^{N \times 1 \times H \times W}
    \end{align}
    
    \State Calculate threshold value $Q$ with given parameter quantile $q$: 
    \begin{align}
        Q = Threshold({[F_{ij}]_{H \times W},q)}, \quad Q \in \mathbb{R}^{N}
    \end{align}
    
    \State Generate input Mask for every sample:
    \begin{align}
    	Mask=[\mathbb{I}_{|F_{ij}|>Q}]_{H \times W}, Mask \in \{0,1\}^{N \times 1 \times H \times W}
    \end{align}
    
    \State Random sampling $r\%$ images from the input, then apply Gaussian blur on their inhibited areas (i.e where mask[i,j]= 0) to obtain the enhanced image.
    \end{algorithmic} 
\end{algorithm} 

\begin{figure}[htb]
\centering
\begin{minipage}[]{0.45\textwidth}
\flushleft 
  \includegraphics[width=\textwidth]{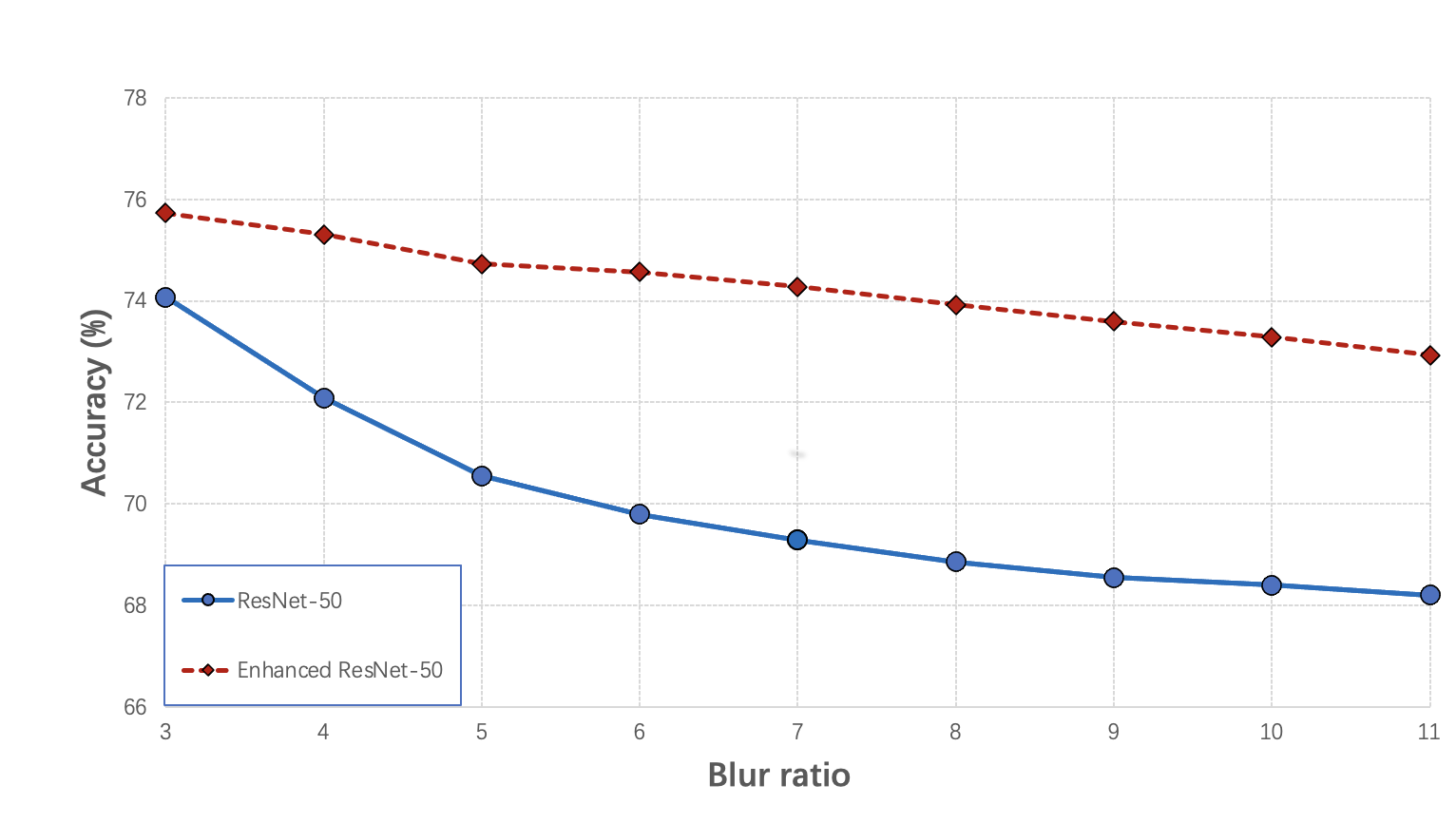}
  \caption{Accuracy of Res-Net50 and enhanced ResNet-50 in 100,000 images with different blur ratio applying on their "background"}
  \label{enhancement}
\end{minipage}
\end{figure}

 Figure \ref{enhancement} shows the result of ResNet-50 and enhanced ResNet-50 on the validation set with different blur ratio applied to their "background". It can be seen that the data-enhanced network has better robustness in the dataset with blurred background.

\section{Conclusion and Discussion}

In this work, we propose a novel CNN training technique inspired by biological lateral inhibition - Gradient Mask - for filtering noise gradients during the backpropagation process so as to improve its performance. The experimental results show that, accuracy of the original CNN architecture, after pruning, and under adversarial attacks, improves. CNN trained with Gradient Mask also generates better saliency maps, which is used for data enhancement which achieves better network interpretability. Furthermore, we provide an analytical explanation for these improvements based on a new criterion for gradient quality: the gradient flux sensitivity.

For future work, we would consider making some of the hyper-parameters, such as the inhibition rate of each layer/bottleneck, $\sigma$ in LoG, the number of groups of channel division, learnable. Gradient Mask applies the notion of LI during back-propagation. How would one implement LI in the forward-propagation, and what purpose would it serve? With the recent advances made by Transformers in CV and NLP \cite{vaswani2017attention,devlin2018bert,brown2020language,dosovitskiy2020image,zhu2020deformable,radford2021learning}, and its potential to unify different neural network architectures \cite{radford2021learning}, and to form minicolumn naturally, one may contemplate how LI maybe introduce to the transformer.

{\small
\bibliographystyle{ieee_fullname}
\bibliography{main}
}

\end{document}